\documentclass[journal]{IEEEtran}
\usepackage{times,amsmath,bm,epsfig}
\usepackage{graphicx}
\usepackage[numbers]{natbib}
\usepackage{verbatim}
\usepackage{array}
\usepackage{color}
\usepackage{tabularx,multirow}
\usepackage{threeparttable}

\newcommand{\eqnref}[1]{(\ref{#1})}
\newcommand{\figref}[1]{Fig.~\ref{#1}}
\newcommand{\secref}[1]{Section~\ref{#1}}
\newcommand{\tabref}[1]{Table~\ref{#1}}

\usepackage{color,soul}
\soulregister\cite7
\soulregister\eqnref7
\soulregister\figref7
\soulregister\secref7
\soulregister\tabref7
\usepackage{xcolor}
\usepackage{multirow}
\begin{document}

\title{DualConv: Dual Convolutional Kernels for Lightweight Deep Neural Networks}

\author{Jiachen~Zhong,~Junying~Chen,~\IEEEmembership{Member,~IEEE,}~and~Ajmal~Mian,~\IEEEmembership{Senior Member,~IEEE}
\thanks{This work was supported in part by the National Natural Science Foundation of China under Grant 61802130, and in part by the Guangdong Natural Science Foundation under Grant 2019A1515012152 and 2021A1515012651. 
Ajmal Mian is the recipient of an Australian Research Council Future Fellowship (project number FT210100268) funded by the Australian Government.
 \textit{(Corresponding author: Junying Chen.)}}
\thanks{J. Zhong and J. Chen are with the School of Software Engineering, South China University of Technology, and also with the Key Laboratory of Big Data and Intelligent Robot (SCUT), Ministry of Education, Guangzhou 510006, China (e-mail: 1173992770@qq.com; jychense@scut.edu.cn).}
\thanks{A. Mian is with the Department of Computer Science, 
The University of Western Australia, Perth, Australia (e-mail: ajmal.mian@uwa.edu.au).}
\thanks{Digital Object Identifier 10.1109/TNNLS.2022.3151138}
\thanks{This work is licensed under a Creative Commons Attribution 4.0 License.}
\thanks{For more information, see https://creativecommons.org/licenses/by/4.0/}
}

\markboth{IEEE TRANSACTIONS ON NEURAL NETWORKS AND LEARNING SYSTEMS}%
{Shell \MakeLowercase{\textit{Zhong~et~al.}}: DualConv: Dual Convolutional Kernels for Lightweight Deep Neural Networks}

\IEEEaftertitletext{\vspace{-1\baselineskip}} 

\maketitle

\begin{abstract}
CNN architectures are generally heavy on memory and computational requirements which makes them infeasible for embedded systems with limited hardware resources. We propose dual convolutional kernels (DualConv) for constructing lightweight deep neural networks. 
DualConv combines 3$\times$3 and 1$\times$1 convolutional kernels to process the same input feature map channels simultaneously and exploits the group convolution technique to efficiently arrange convolutional filters. 
DualConv can be employed in any CNN model such as VGG-16 and ResNet-50 for image classification, YOLO and R-CNN for object detection, or FCN for semantic segmentation. In this paper, we extensively test DualConv for classification since these network architectures form the backbones for many other tasks. We also test DualConv for image detection on YOLO-V3.
Experimental results show that, combined with our structural innovations, DualConv significantly reduces the computational cost and number of parameters of deep neural networks while surprisingly achieving slightly higher accuracy than the original models in some cases. 
We use DualConv to further reduce the number of parameters of the lightweight MobileNetV2 by 54\% with only 0.68\% drop in accuracy on CIFAR-100 dataset. When the number of parameters is not an issue, DualConv increases the accuracy of MobileNetV1 by 4.11\% on the same dataset.
Furthermore, DualConv significantly improves the YOLO-V3 object detection speed and improves its accuracy by 4.4\% on PASCAL VOC dataset.
\end{abstract}


\section{Introduction}
\IEEEPARstart{C}{ONVOLUTIONAL} Neural Networks (CNNs) have achieved unmatched performance in many applications such as image classification, object detection, and semantic segmentation. Current research trend of improving and enhancing network performance makes the networks deeper and more complex, which eventually leads to a dramatic increase in the model size (number of parameters/weights) and the required computational resources. Due to these two reasons, modern CNN models can only run on servers equipped with high-performance GPUs. Although embedded devices and mobile platforms have a huge demand for deployment of deep models, current architectures are not suitable for these systems due to their limited memory, power and computational resources. Therefore, designing lightweight yet accurate CNN models that can be deployed in embedded devices and mobile platforms, has become an active research direction.

In embedded devices and mobile platforms, the network accuracy, computational complexity and number of parameters are all equally important factors for evaluating different network architectures. Hence, many methods have been proposed to increase the efficiency of neural network models. A general approach taken by these methods is to start from a standard CNN model and increase the model efficiency by reducing the number of parameters and floating-point operations (FLOPs) through model compression. Model compression can be divided into three broad categories: connection pruning~\cite{han2016deep}, filter pruning~\cite{yu2018nisp,lin2017runtime,luo2017thinet,he2017channel,ding2018auto,li2017pruning,neklyudov2017structured} and model quantization~\cite{han2016deep,rastegari2016xnor}.

Network connection pruning not only reduces the network complexity, but also prevents the network from over-fitting. Hanson \textit{et al.} proposed a pruning method based on bias parameter attenuation~\cite{hanson1989comparing}. LeCun \textit{et al.} showed that a trade-off can be made between network accuracy and complexity by using second-derivative information to remove unimportant weights from the network~\cite{lecun1990optimal}. Hassibi \textit{et al.} further extended the idea and argued that retraining a highly pruned network (as in~\cite{lecun1990optimal}) may lead to inferior generalization~\cite{hassibi1993second}. However,~\cite{lecun1990optimal} and~\cite{hassibi1993second} are both based on the computation of the Hessian matrix which incurs a high computational cost. The idea of filter pruning is to prune the filter channels which contribute the least in the network model~\cite{luo2017thinet,he2017channel}. Lin \textit{et al.} modeled network channel-wise pruning as a Markov decision process and used reinforcement learning for training~\cite{lin2017runtime}. They named this method as runtime neural pruning (RNP) as it pruned the deep neural network dynamically at runtime. Luo \textit{et al.}~\cite{luo2017thinet} and Li \textit{et al.}~\cite{li2017pruning} focused on filter level pruning which pruned the whole filter if it was less important. He \textit{et al.} used LASSO regression to select filter channels and least square reconstruction to rebuild the network~\cite{he2017channel}. After pruning the network, the model usually needs fine-tuning to maintain its performance~\cite{yu2018nisp}. In model quantization, the idea is to reduce the parameters of the network or to reduce the storage bits of the feature maps~\cite{han2016deep,rastegari2016xnor}. For example, Vanhoucke \textit{et al.} used 8-bit unsigned char to reduce the storage bits of the activation values~\cite{vanhoucke2011improving}. The reasoning behind this is that the accuracy/precision of the weights in the process of network inference does not need to be so high.

In practice, model compression is a costly and difficult process. 
Therefore, there is a need to design efficient networks right from the start. 
One such method is to design efficient network architectures that inherently have fewer parameters and lower complexity. An efficiently designed network architecture requires less data and time to train and is also easy to prune after training. However, designing a new network architecture that maintains high accuracy with minimal computational cost requires significant effort given the large number and space of hyperparameters. Moreover, this approach also does not take advantage of the many existing standard network architectures.
Hence, a better approach is to design efficient convolutional filters for existing standard network architectures. The new convolutional filters can simply replace standard convolutional filters in existing CNN model architectures to reduce their computational complexity without sacrificing accuracy. 

SqueezeNet~\cite{iandola2016squeezenet} proposed by Han \textit{et al.} significantly reduces parameters and computational complexity while maintaining network accuracy. Its network structure unit introduces 1$\times$1 convolution to reduce the computational complexity. The 1$\times$1 convolutional kernels not only reduce the network parameters and computational complexity, but also offer the flexibility to control the depth of the feature maps, achieve cross-channel information fusion, and provide an additional level of non-linearity. Depthwise separable convolution was proposed in MobileNetV1~\cite{howard2017mobilenets} by Howard \textit{et al.} The depthwise separable convolution is a form of factorized convolution that factorizes a standard convolution into a depthwise and a 1$\times$1 pointwise convolution. When the 1$\times$1 convolution is used to filter the input feature maps, it fuses the original information of each channel of input feature maps into the output feature maps. Hence, the original information of input images can pass to deeper convolutional layers. MobileNetV2~\cite{sandler2018mobilenetv2} proposed inverted residual blocks which first use 1$\times$1 convolutions to increase the number of channels of the input feature maps and then use depthwise separable convolutions to filter the features. 
ShuffleNet~\cite{zhang2018shufflenet} proposed by Zhang \textit{et al.} uses a form of convolution called group convolution (GroupConv) to reduce the computational cost of the network. It also uses the channel shuffle operation to enhance the interaction between different groups of channels. Heterogeneous convolution (HetConv)~\cite{singh2019hetconv} uses heterogeneous convolutional kernels with different sizes within a convolutional filter. 
Whereas the 3$\times$3 convolution in HetConv extracts the spatial information of input feature map, the 1$\times$1 convolution in HetConv reduces the computational cost of neural network allowing for information sharing between convolutional layers. Heterogeneous convolutional filters are applicable to existing standard network architectures to reduce the network complexity.

Inspired by GroupConv and HetConv, we propose dual convolution (DualConv), designing a new convolutional filter which integrates 3$\times$3 group convolution with 1$\times$1 pointwise convolution to deal with the same input feature map channels simultaneously. Because DualConv uses 1$\times$1 convolution to preserve the original information of input feature maps, the 3$\times$3 convolutional filters at deeper convolutional layers can still learn from the original information of input feature maps. DualConv is more efficient and more general compared to model compression methods, because it can be applied to all current and future CNN architectures. 

The proposed DualConv is used to replace the standard convolution in VGG-16~\cite{simonyan2014very} and ResNet-50~\cite{he2016deep} to perform image classification experiments on CIFAR-10~\cite{cifar}, CIFAR-100~\cite{cifar} and the large-scale ImageNet~\cite{russakovsky2015imagenet} datasets. Our results show that the proposed DualConv significantly reduces the cost of network computation and the number of parameters while surprisingly achieving slightly higher accuracy than the original models in some cases. For better comparison, we reproduce GroupConv and HetConv to replace the standard convolution in VGG-16 on CIFAR-10 dataset, and replace the standard convolution in ResNet-50 on ImageNet dataset. Quantitative results and visual analysis show that DualConv generally achieves higher accuracy than GroupConv and HetConv with slightly higher number of parameters. DualConv is further applied to modify the convolutional filters in MobileNetV1. Our DualConv-modified MobileNetV1 performs more accurately than the original MobileNetV1 as well as the GroupConv-modified (or HetConv-modified) MobileNetV1. DualConv is also applied to modify the convolutional filters in MobileNetV2, reducing its parameters by 54\% with only 0.68\% drop in accuracy on CIFAR-100 dataset. 

The proposed DualConv is further tested on object detection task by replacing the 3$\times$3 standard convolution in YOLO-V3~\cite{redmon2018yolov3}. Experiments on PASCAL VOC dataset~\cite{Everingham10} show that DualConv-modified YOLO-V3 requires much less computations leading to faster detection speed. Moreover, DualConv-modified YOLO-V3 improves the mAP value of each image with 4.4\% higher accuracy.

\section{Related Work}
Efficient convolutional filters can effectively reduce the computational cost and parameters of a neural network, and eliminate the need for designing new convolutional network architectures from scratch. Three types of efficient convolutional filters have been proposed in the literature to replace standard convolutional filters in existing network architectures. We briefly describe these below.

\begin{figure*}[t]
\centering
\includegraphics[width=0.9\textwidth]{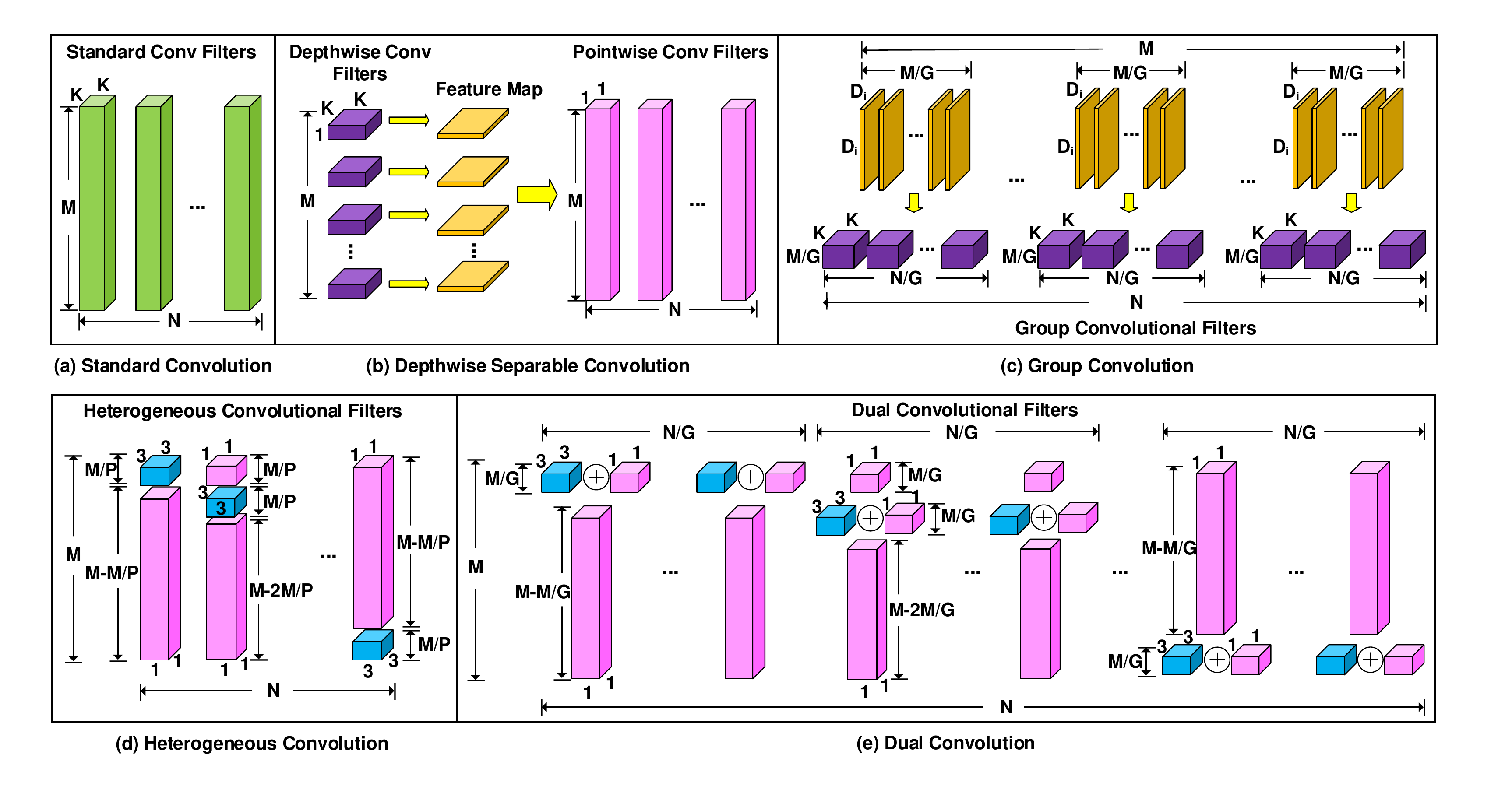} 
\caption{Convolutional filter designs of (a) standard convolution, (b) depthwise separable convolution, (c) group convolution, (d) heterogeneous convolution, and (e) the proposed dual convolution. $M$ is the number of input channels (\textit{i.e.}, the depth of input feature map), $N$ is the number of convolutional filters and also the number of output channels (\textit{i.e.}, the depth of output feature map), $D_i$ is the width and height dimension of input feature map, $K \times K$ is the convolutional kernel size, $G$ is the number of groups in group convolution and dual convolution, and $1/P$ is the ratio of 3$\times$3 convolutional kernels in heterogeneous convolution. Note that the heterogeneous filters are arranged in a shifted manner~\cite{singh2019hetconv}.}
\label{fig1}
\end{figure*}

\subsection{Depthwise Separable Convolution}
A standard convolution, shown in \figref{fig1}(a), simultaneously performs feature extraction and channel fusion on the input feature maps. Depthwise separable convolution in MobileNetV1~\cite{howard2017mobilenets} decomposes the standard convolution into depthwise convolution and pointwise convolution as shown in \figref{fig1}(b). In depthwise convolution, a single convolutional kernel is applied to each input channel. Usually, a 3$\times$3 convolution is used for such feature extraction. Pointwise convolution applies 1$\times$1 convolution to the output feature map of depthwise convolution to perform channel-wise fusion. Hence, by splitting the feature extraction and channel fusion, the depthwise separable convolution significantly reduces the number of parameters and consequently the computations performed by the network.

\subsection{Group Convolution}
The concept of GroupConv was first proposed in AlexNet~\cite{krizhevsky2012imagenet}. Due to limited GPU performance, at that time, the model was divided into two GPUs for training. In GroupConv, the convolutional filters are divided into $G$ groups and the input feature map channels are also divided into $G$ groups as shown in \figref{fig1}(c). Each group of convolutional filters processes the corresponding group of input feature map channels. Since each group of convolutional filters is only applied to the corresponding input channel group, the computational cost of convolution is significantly reduced, but the channel information is not shared between different groups, \textit{i.e.}, different groups of output feature map channels only receive information from their corresponding groups of input channels. This hinders the flow of information between different groups of channels, reducing the feature extraction ability of GroupConv. To overcome this issue, ShuffleNet~\cite{zhang2018shufflenet} performs a channel shuffle operation to enhance the information exchange between different groups of channels.

\begin{figure}[t]
\centering
\includegraphics[width=0.98\columnwidth]{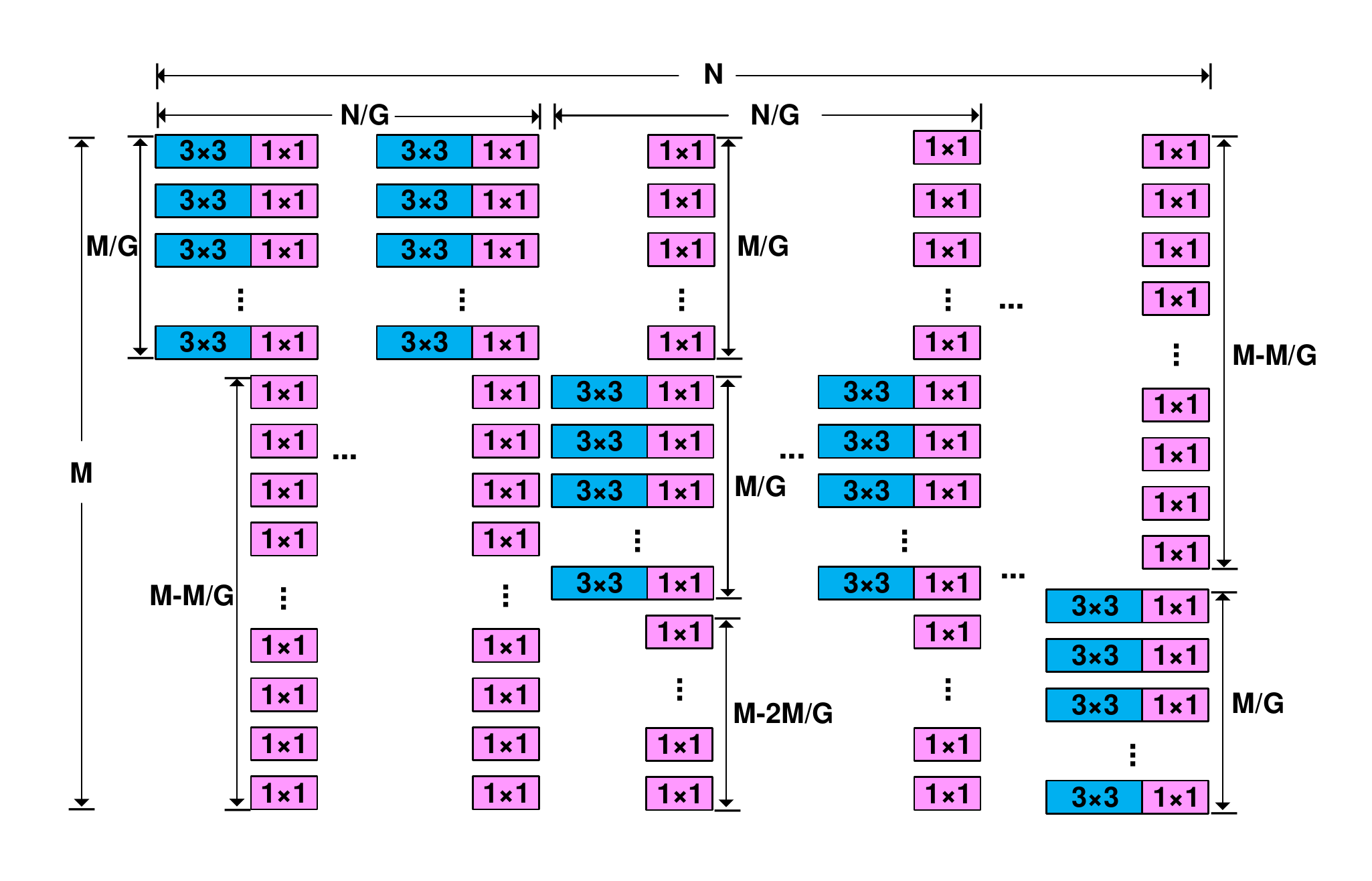} 
\caption{Structural layout of dual convolution.}
\label{fig2}
\end{figure}

\subsection{Heterogeneous Convolution}
HetConv~\cite{singh2019hetconv} contains both 3$\times$3 convolution and 1$\times$1 convolution in one convolutional filter, as shown in \figref{fig1}(d). Note that the heterogeneous filters are arranged in a shifted manner (see Fig. 3 in~\cite{singh2019hetconv}). The $M/P$ 3$\times$3 convolutional kernels are discretely arranged, and the 3$\times$3 and 1$\times$1 kernels alternate within a convolutional filter.
The computational complexity of the original 3$\times$3 standard convolution can be reduced by three to eight times using heterogeneous convolutional filters, without sacrificing the accuracy of the network much. 
The heterogeneous design essentially breaks down the continuity of cross-channel information integration and negatively affects the preservation of complete information of input feature map. Therefore, such a strategy could reduce the network accuracy.

\section{The Proposed Dual Convolution}

\subsection{Design Scheme of Dual Convolution}
\label{dualconv}
We propose dual convolution which combines the strengths of group convolution and heterogeneous convolution. Whereas some convolutional kernels perform both 3$\times$3 and 1$\times$1 convolutional operations simultaneously, others only perform 1$\times$1 convolutions, as shown in \figref{fig1}(e). The structural layout of dual convolution is shown in \figref{fig2}. Notice how the 3$\times$3 convolution moves in the feature map channel dimension and yet the 1$\times$1 convolution is performed on all input channels. Our approach can be regarded as the combination of 3$\times$3 group convolution and 1$\times$1 pointwise convolution on the same input feature map, which makes it easy to integrate into existing network architectures. Because applying continuous 1$\times$1 convolution on input feature maps can preserve the original information, it can help deeper convolutional layers to extract information more effectively.

DualConv not only solves the problem of poor communication of GroupConv, but also improves the performance of deep neural networks compared to HetConv. In the GroupConv shown in \figref{fig1}(c), 
every $N/G$ convolutional filters handle $M/G$ input feature map channels, extracting information for $N/G$ output feature map channels. As each convolutional filter extracts information from only $1/G$ input channels, the output feature map channel of such convolutional filter contains less information than that of the convolutional filter which handles the complete input feature map. Based on this observation, we add $M$ 1$\times$1 convolutional kernels to each convolutional filter so that it is able to handle the complete input feature map for better information extraction and sharing between convolutional layers. In the HetConv shown in \figref{fig1}(d), $M/P$ kernels are 3$\times$3 convolutional kernels, and the rest ($M-M/P$) kernels are 1$\times$1 convolutional kernels. Such alternative arrangement breaks down the continuity of cross-channel information integration and negatively affects the preservation of complete information of input feature map. Based on this observation, we design parallel 1$\times$1 convolutional kernels for all the 3$\times$3 convolutional kernels, so as to preserve the original information of input feature maps to help deeper convolutional layers to extract more effective features.

We combine the above two modifications together to design DualConv. We divide $N$ convolutional filters into $G$ groups, each group handles the complete input feature map where $M/G$ input feature map channels are processed by 3$\times$3 and 1$\times$1 convolutional kernels simultaneously and the rest ($M-M/G$) input channels are processed by 1$\times$1 convolutional kernels solely. The results of simultaneous 3$\times$3 and 1$\times$1 convolutional kernels are summed up, as indicated by the $\oplus$ sign in \figref{fig1}(e). 
Because the filter group structure enforces a block-diagonal sparsity on the channel dimension, the filters with high correlation are learned in a more structured way~\cite{yani2017deeproots}. As such, we do not arrange the convolutional filters in a shifted manner.
The design of DualConv reduces the parameters of original backbone network models through group convolution strategy, and promotes better information sharing between convolutional layers by preserving the original information of input feature maps and allowing for maximum cross-channel communication with $M$ 1$\times$1 convolutions. As a result, DualConv can be constructed without the need for channel shuffle operation.

Assume that the size of output feature map is $D_{o} \times D_{o} \times N$, where $D_{o}$ is the width and height dimension of output feature map. In the standard convolution shown in \figref{fig1}(a), the input feature map is filtered by $N$ convolutional filters with size of $K \times K \times M $ in the convolutional layer, where $K \times K$ is the convolutional kernel size. Therefore, the total number of FLOPs performed in a standard convolutional layer $FL_{SC}$ is:
\begin{equation}
    FL_{SC} = D_{o}^2 \times K^2 \times M \times N.
\end{equation}

In DualConv, the number of convolutional filter groups $G$ is used to control the proportion of $K \times K$ convolutional kernels in a convolutional filter. For a given $G$, the proportion of combined simultaneous convolutional kernels with size of ($K \times K + 1 \times 1$) is $1/G$ of all channels, while the proportion of the remaining 1$\times$1 convolutional kernels is ($1-1/G$). Therefore, in a dual convolutional layer composed of $G$ convolutional filter groups, the number of FLOPs for the combined convolutional kernels is:
\begin{equation}
    FL_{CC} = (D_{o}^2 \times K^2 \times M \times N + D_{o}^2 \times M \times N ) / G,
\end{equation}
and the number of FLOPs for the remaining 1$\times$1 pointwise convolutional kernels is:
\begin{equation}
    FL_{PC} = (D_{o}^2 \times M \times N) \times (1 - 1/G).
\end{equation}
The total number of FLOPs is:
\begin{equation}
\begin{aligned}
    FL_{DC} &= FL_{CC} + FL_{PC}\\&=D_{o}^2 \times K^2 \times M \times N/G + D_{o}^2 \times M \times N.
\end{aligned}
\end{equation}

Comparing the computational cost (FLOPs) of dual convolutional layer with that of standard convolutional layer, the computational reduction ratio $R_{DC/SC}$ is:
\begin{equation}
    R_{DC/SC} = \frac{FL_{DC}}{FL_{SC}} = \frac{1}{G}+\frac{1}{K^2}.
\label{eqn_rdc}
\end{equation}
As seen from \eqnref{eqn_rdc}, given that $K$=3 in DualConv design, the speedup can reach 8 to 9 times when $G$ is large.

\subsection{Comparison with Previous Work}
As shown in \figref{fig1}(b), a depthwise separable convolutional layer contains two convolutional layers, \textit{i.e.}, a depthwise convolutional layer followed by a pointwise convolutional layer, which increases the network complexity. On the contrary, the proposed DualConv does not add additional layers to the network. The number of FLOPs for a depthwise separable convolutional layer is:
\begin{equation}
    FL_{DSC} = D_{o}^2 \times( K^2 \times M + M \times N ).
\end{equation} 
The computational reduction ratio over the standard convolutional layer $R_{DSC/SC}$ is: 
\begin{equation}
    R_{DSC/SC} = \frac{FL_{DSC}}{FL_{SC}} = \frac{1}{N} + \frac{1}{K^{2}}.
\label{eqn_rdsc}
\end{equation} 
As mentioned in \secref{dualconv}, each convolutional filter in GroupConv extracts information from only $1/G$ input channels, while the convolutional filter in DualConv handles the complete input feature map. In a group convolutional layer, the number of FLOPs is:
\begin{equation}
    FL_{GC} = (D_{o}^2 \times K^2 \times M \times N) / G,
\end{equation} 
and the computational reduction ratio $R_{GC/SC}$ is:
\begin{equation}
    R_{GC/SC} = \frac{FL_{GC}}{FL_{SC}} = \frac{1}{G}.
\label{eqn_rgc}
\end{equation} 

Unlike HetConv where the 1$\times$1 convolution is not applied to all input feature map channels, the proposed DualConv operates 1$\times$1 convolution on the whole input feature map. It can retain and fuse the information of the original input features better than HetConv, with only a slight increase in the number of FLOPs and parameters. The number of FLOPs for a heterogeneous convolutional layer is:
\begin{equation}
    FL_{HC} = (D_{o}^2 \times M \times N ) \times \frac{K^2 + P - 1}{P}, 
\end{equation} 
and the computational reduction ratio $R_{HC/SC}$ is:
\begin{equation}
    R_{HC/SC} = \frac{FL_{HC}}{FL_{SC}} = \frac{1}{P} + \frac{1}{K^{2}} - \frac{1}{P \times K^{2}}.
\label{eqn_rhc}
\end{equation} 

As derived from \eqnref{eqn_rdsc} and \eqnref{eqn_rhc}, given $K$=3, when $N$ and $P$ are large, the speedup of depthwise separable convolution and HetConv can reach 8 to 9 times, which is similar to the speedup of DualConv. However, the speedup of GroupConv is proportional to $G$.

\section{Experiments and Discussions}
We perform extensive experiments using the proposed dual convolutional filters. 
The trade-off between accuracy and computational cost of the network model is adjusted by the number of convolutional filter groups, $G$. When the value of $G$ is large, the structure of DualConv becomes closer to the standard convolution consisting of all 1$\times$1 convolutional kernels. In general, DualConv retains the accuracy of the original network and in some cases achieves slightly higher accuracy than the original convolutional filter. Moreover, compared to other efficient convolutions with similar computational cost, DualConv achieves higher network accuracy. Hence, DualConv makes it more feasible to deploy deep CNNs on mobile platforms or embedded devices.



\subsection{VGG-16 and ResNet-50 on CIFAR-10}
\label{std_cifar10}
For VGG-16 network architecture, we replace the 3$\times$3 standard convolutions in the last 12 layers with the proposed DualConv. The $G$ values for all replaced layers are the same, and the number of convolutional kernels in each layer is kept the same as that of the original VGG-16 network. In ResNet-50 network structure, we use DualConv to replace all the 3$\times$3 standard convolutions with stride 1 in the convolutional layers (except the first layer). For hyperparameters, we set the weight decay to 5e-4, the initial learning rate to 0.1 and multiply it by 0.1 after every 50 epochs. We use SGD optimizer and the multiply step learning rate decay strategy.

\begin{table}[t]
\caption{Performance of VGG-16 and ResNet-50 with DualConv, GroupConv or HetConv on CIFAR-10 using different settings\textsuperscript{1}.}
\footnotesize
\begin{center}
\begin{tabular}{|l|c|c|c|c|c|}
\hline
Model&Acc (\%)&FLOPs&Params\\
\hline\hline
VGG-16&93.95&313.21M&14.73M\\
\hline
Li-pruned~\cite{li2017pruning}&93.40&206.44M&5.32M\\
SBP~\cite{neklyudov2017structured}&92.50&136.41M&-\\
SBPa~\cite{neklyudov2017structured}&91.00&99.30M&-\\
AFP-E~\cite{ding2018auto}&92.94&63.72M&-\\
AFP-F~\cite{ding2018auto}&92.87&58.39M&-\\
\hline
VGG-16\_GC\_G2&92.48&157.49M&7.37M\\
VGG-16\_GC\_G4&90.23&79.64M&3.69M\\
VGG-16\_GC\_G8&88.46&40.71M&1.85M\\
VGG-16\_GC\_G16&85.78&21.24M&0.93M\\
VGG-16\_GC\_G32&82.62&11.51M&0.48M\\
\hline
VGG-16\_HC\_P2&93.69&175.23M&8.45M\\
VGG-16\_HC\_P4&93.67&105.98M&5.17M\\
VGG-16\_HC\_P8&93.52&71.35M&3.54M\\
VGG-16\_HC\_P16&93.34&54.04M&2.72M\\
VGG-16\_HC\_P32&92.97&45.38M&2.31M\\
\hline
VGG-16\_G2&93.91&192.10M&9.00M\\
VGG-16\_G4&\textbf{94.14} &114.24M&5.33M\\
VGG-16\_G8&93.61&75.31M&3.49M\\
VGG-16\_G16&93.55&55.85M&2.57M\\
VGG-16\_G32&93.20&46.11M&2.11M\\
\hline\hline
ResNet-50&94.08&1.30G&23.52M\\
\hline
ResNet-50\_G2&93.81&1.11G&20.32M\\
ResNet-50\_G4&94.15 &984.33M&18.27M\\
ResNet-50\_G8&\textbf{94.33}&922.99M&17.24M\\
ResNet-50\_G16&93.68&892.32M&16.73M\\
ResNet-50\_G32&93.08&876.98M&16.47M\\
\hline
\end{tabular}
\begin{tablenotes}
 \item[] \textsuperscript{1}VGG-16(ResNet-50)\_G$\alpha$ represents DualConv-modified VGG-16 or ResNet-50, VGG-16\_GC\_G$\alpha$ represents GroupConv-modified VGG-16, and VGG-16\_HC\_P$\alpha$ represents HetConv-modified VGG-16, where $\alpha$ is the value of $G$ in DualConv and GroupConv, or the value of $P$ in HetConv.
\end{tablenotes}
\end{center}
\label{table1}
\end{table}



\tabref{table1} shows the performance comparisons of DualConv, GroupConv and HetConv on CIFAR-10 dataset using VGG-16 network architecture. \tabref{table1} also illustrates the comparisons with several representative model compression methods applied to VGG-16, \textit{e.g.}, Li-pruned~\cite{li2017pruning}, SBP~\cite{neklyudov2017structured} and AFP~\cite{ding2018auto}. For a fair comparison between different convolutional filters under the same implementation framework, the proposed DualConv and the reproduced GroupConv and HetConv are all implemented in the PyTorch framework adopting the im2col method to flatten the feature maps and convolutional kernels. The implementation framework and the flattening method may be the reasons why the reproduced HetConv performs slightly (0.2$\sim$0.76\%) worse than the HetConv reported in~\cite{singh2019hetconv}. 

In \tabref{table1}, with the increase of $G$ value, the number of FLOPs and parameters of network decrease significantly, whereas the network accuracy drops slightly. When $G$=4, the accuracy of VGG-16 network is actually higher than that of standard VGG-16 network while the computations and the parameters are both reduced by over 60\%. DualConv generally obtains higher accuracy than HetConv, demonstrating better feature learning ability than HetConv, since $1\times 1$ convolution is applied to all channels. Moreover, as for ResNet-50, the accuracy of ResNet-50 with DualConv outperforms the standard ResNet-50 network and the computations and parameters are reduced significantly by over 25\% when $G$=8.


Our results demonstrate that 1$\times$1 pointwise convolution can transfer and fuse the information of input feature maps well, and the output feature maps can retain the information of input feature maps better. Hence, it can be performed without the need for channel shuffle operation.

\subsection{MobileNetV1 and MobileNetV2 on CIFAR-10}

\begin{table}[t]
\caption{Performance of MobileNetV1 and MobileNetV2 with DualConv on CIFAR-10 using different settings.}
\footnotesize
\begin{center}
\begin{tabular}{|l|c|c|c|c|c|}
\hline
Model&Acc (\%)&FLOPs&Params\\
\hline\hline
MobileNetV1&91.91&46.37M&3.22M\\
\hline
MobileNetV1\_HC\_P32~\cite{singh2019hetconv}&92.17&56.91M&-\\
\hline
MobileNetV1\_G2&93.09&243.12M&17.29M\\
MobileNetV1\_G4&\textbf{93.14}&144.03M&10.23M\\
MobileNetV1\_G8&92.92&94.49M&6.69M\\
MobileNetV1\_G16&92.64&69.71M&4.93M\\
MobileNetV1\_G32&91.92&57.3M&4.04M\\
\hline\hline
MobileNetV2&91.99&64.96M&2.37M\\
\hline
MobileNetV2\_G2&90.71&41.84M&1.45M\\
MobileNetV2\_G4&90.56&31.07M&1.09M\\
MobileNetV2\_G8&90.56&25.68M&916.92K\\
MobileNetV2\_G16&90.31&23.50M&829.94K\\
MobileNetV2\_G32&\textbf{90.83}&22.42M&786.45K\\
\hline
\end{tabular}
\end{center}
\label{table3}
\end{table}

In MobileNetV1 network architecture, we replace all the depthwise separable convolutions with our proposed dual convolutions. Since the images in CIFAR-10 and CIFAR-100 datasets are much smaller than the images in ImageNet dataset, the stride of the first depthwise separable convolutional layer in the original or DualConv-modified MobileNetV1 network is modified to 1 instead of 2. In MobileNetV2 network structure, we replace the inverted residual block with our proposed dual convolution while the convolution stride is kept as 1. The replacement strategy is to add batch normalization and ReLU6 operations after the proposed dual convolution when replacing the inverted residual block.
The experimental settings are the same as those in \secref{std_cifar10}.

\tabref{table3} shows our experimental results. 
In MobileNetV1 network structure, although the proposed DualConv increases the parameters and computational cost of the original MobileNetV1 network, it improves the network accuracy by 1.23\%. Even when $G$=32, MobileNetV1 with DualConv still has higher accuracy than the original MobileNetV1 network with similar parameters and computational cost. In MobileNetV2 network structure, our proposed DualConv can decrease the network parameters and computational cost exceeding 60\% with only 1.16\% drop in accuracy when $G$=32.

\subsection{VGG-16 and ResNet-50 on CIFAR-100}
\label{std_cifar100}
In this experiment, we use DualConv to modify VGG-16 and ResNet-50 network structures to perform image classification on a larger dataset CIFAR-100. We replace the 3$\times$3 standard convolutions with dual convolutions. Note that, the VGG-16 architecture used for CIFAR-100 dataset has three fully connected layers, while the VGG-16 architecture used for CIFAR-10 dataset has only one fully connected layer. The values of the hyperparameters are set as: weight decay=5e-4, and initial learning rate=0.1 which is multiplied by 0.2 after every 60 epochs. Moreover, we use SGD optimizer and multiply step learning rate decay strategy.

The Top-1 accuracy of the networks are recorded in \tabref{table4}. As shown in \tabref{table4}, when we replace the 3$\times$3 standard convolutions of VGG-16 with the proposed dual convolutions, the accuracy improves when $G$ increases to 8. The computational cost is reduced by more than 70\% when $G$=8. When $G$ increases further, the accuracy of network drops slightly, but the number of parameters and computational cost are further reduced. On the other hand, the best accuracy of ResNet-50 with DualConv is achieved when $G$=4, which is only 0.02\% lower than the original ResNet-50, but the number of parameters and computational cost are reduced by more than 20\%. These results demonstrate the strong generalization ability of the proposed DualConv.

\begin{table}[t]
\caption{Performance of VGG-16 and ResNet-50 with DualConv on CIFAR-100 using different settings.}
\footnotesize
\begin{center}
\begin{tabular}{|l|c|c|c|c|c|}
\hline
Model&Acc (\%)&FLOPs&Params\\
\hline\hline
VGG-16&72.41&332.48M&34.01M\\
\hline
VGG-16\_G2&\textbf{73.04}&211.37M&28.29M\\
VGG-16\_G4&72.82&133.52M&24.61M\\
VGG-16\_G8&72.52&94.59M&22.78M\\
VGG-16\_G16&72.07&75.12M&21.86M\\
VGG-16\_G32&71.52&65.39M&21.40M\\
\hline\hline
ResNet-50&78.55&1.30G &23.70M\\
\hline
ResNet-50\_G2&78.46&1.11G&20.50M\\
ResNet-50\_G4&\textbf{78.53}&984.51M&18.45M\\
ResNet-50\_G8&78.01&923.17M&17.43M\\
ResNet-50\_G16&77.77&892.50M&16.91M\\
ResNet-50\_G32&77.50&877.17M&16.65M\\
\hline
\end{tabular}
\end{center}
\label{table4}
\end{table}

\subsection{MobileNetV1 and MobileNetV2 on CIFAR-100}
In this experiment, we test the proposed DualConv in MobileNetV1 and MobileNetV2 network structures to perform image classification on CIFAR-100 dataset. The experimental settings are the same as those in \secref{std_cifar100}.

As shown in \tabref{table5}, when we replace the depthwise separable convolutions of MobileNetV1 network with the proposed dual convolutions, the accuracy increases by 4.11\% when $G$=4. Even when the $G$ value increases to 32, the network accuracy still outperforms the standard MobileNetV1 architecture and has similar computational cost. In MobileNetV2 network structure, the proposed dual convolution greatly reduces the network parameters and computational cost with only a slight drop in accuracy (0.68\% when $G$=2 or $G$=4). These experimental results also confirm the strong generalization ability of the proposed DualConv.

\begin{table}[t]
\caption{Performance of MobileNetV1 and MobileNetV2 with DualConv on CIFAR-100 using different settings.}
\footnotesize
\begin{center}
\begin{tabular}{|l|c|c|c|c|c|}
\hline
Model&Acc (\%)&FLOPs&Params\\
\hline\hline
MobileNetV1&68.13&46.46M&3.32M\\
\hline
MobileNetV1\_G2&72.05&243.21M&17.38M\\
MobileNetV1\_G4&\textbf{72.24}&144.12M&10.32M\\
MobileNetV1\_G8&71.79&94.58M&6.79M\\
MobileNetV1\_G16&71.01&69.81M&5.02M\\
MobileNetV1\_G32&70.28&57.42M&4.14M\\
\hline\hline
MobileNetV2&68.54&64.96M&2.37M\\
\hline
MobileNetV2\_G2&\textbf{67.86}&41.84M&1.45M\\
MobileNetV2\_G4&\textbf{67.86}&31.07M&1.09M\\
MobileNetV2\_G8&67.64&25.68M&916.92K\\
MobileNetV2\_G16&67.19&23.50M&829.94K\\
MobileNetV2\_G32&67.22&22.42M&786.45K\\
\hline
\end{tabular}
\end{center}
\label{table5}
\end{table}

\subsection{VGG-16 and ResNet-50 on ImageNet}
\label{std_imagenet}
We experiment with the DualConv-modified VGG-16 and ResNet-50 network architectures on the large-scale ImageNet dataset. The modification strategy of VGG-16 is the same as that described in \secref{std_cifar10}, and VGG-16 for ImageNet also has three fully connected layers as that for CIFAR-100. For ResNet-50 network, we use DualConv to replace all the 3$\times$3 standard convolutions with stride 1 and 2 in the convolutional layers (except the first layer). Moreover, the values of the hyperparamters are set as: weight decay=1e-4, batch size=128, initial learning rate=0.01 in VGG-16 and 0.1 in ResNet-50, and the learning rate is multiplied by 0.1 after every 30 epochs. Both network architectures are trained for 90 epochs.

\begin{table}[t]
\caption{Performance of VGG-16 and ResNet-50 with DualConv on ImageNet using different settings\textsuperscript{1}.}
\footnotesize
\begin{center}
\setlength{\tabcolsep}{0.6mm}{
\begin{tabular}{|l|c|c|c|c|c|c|}
\hline
Model&Top-1&Top-5&FLOPs&Params&GTime&CTime \\
 &Acc(\%)&Acc(\%)& & &(ms)&(ms) \\
\hline\hline
VGG-16&72.69&90.91&15.47G&138.36M&1.74&307.96\\
\hline
RNP (3X)~\cite{lin2017runtime}&-&87.57&5.16G&-&-&-\\
ThiNet-70~\cite{luo2017thinet}&69.8&89.53&4.79G&131.44M&76.71\textsuperscript{*}&-\\
CP 2X~\cite{he2017channel}&-&89.90&7.74G&-&-&-\\
\hline
VGG-16\_HC\_P4~\cite{singh2019hetconv}&71.2&90.20&5.29G&-&-&-\\
\hline
VGG-16\_G2&\textbf{72.21}&\textbf{90.74}&9.54G&132.64M&1.70&239.28\\
VGG-16\_G4&70.74&89.87&5.72G&128.96M&1.54&185.05\\
VGG-16\_G8&69.74&89.18&3.81G&127.13M&1.50&161.86\\
VGG-16\_G16&68.79&88.75&2.86G&126.21M&1.54&\textbf{148.61}\\
VGG-16\_G32&68.22&88.06&2.38G&125.75M&1.52&161.78\\
\hline\hline
ResNet-50&74.27&92.09&4.09G&25.56M&1.34&99.21\\
\hline
ThiNet-50~\cite{luo2017thinet}&71.01&90.02&3.41G&12.38M&153.60\textsuperscript{*}&-\\
NISP~\cite{yu2018nisp}&72.67&-&2.97G&14.36M&-&-\\
\hline
ResNet-50\_GC\_G2&73.85&91.77&3.16G&19.90M&1.22&80.07\\
ResNet-50\_HC\_P4&73.16&91.27&2.86G&18.02M&1.36&92.83\\
\hline
ResNet-50\_G2&\textbf{74.09}&\textbf{91.80}&3.37G&21.16M&1.34&87.46\\
ResNet-50\_G4&74.03&91.76&2.91G&18.33M&1.32&79.47\\
ResNet-50\_G8&73.83&91.77&2.68G&16.91M&1.32&\textbf{75.19}\\
ResNet-50\_G16&73.49&91.31&2.56G&16.20M&1.34&75.88\\
ResNet-50\_G32&72.98&91.05&2.50G&15.85M&1.32&76.10\\
\hline
\end{tabular}}
\begin{tablenotes}
 \item[ ] \textsuperscript{*}Recorded on a single M40 GPU with batch size of 32.
 \item[ ] \textsuperscript{1}GTime represents GPU time, and CTime represents CPU time.
\end{tablenotes}
\end{center}
\label{table6}
\end{table}

\tabref{table6} presents the performance of VGG-16 and ResNet-50 networks with DualConv on ImageNet. Some representative model compression methods (including RNP (3X)~\cite{lin2017runtime}, CP 2X~\cite{he2017channel}, ThiNet~\cite{luo2017thinet} and NISP~\cite{yu2018nisp}), HetConv-modified VGG-16 and ResNet-50 (achieving the best performance when $P$=4), and GroupConv-modified ResNet-50 (achieving the best performance when $G$=2) are also compared.

As illustrated in \tabref{table6}, when $G$=2, the computational cost of VGG-16 with DualConv decreases by about 38\% compared to the original VGG-16 network with only a slight drop in accuracy (0.48\% in Top-1 accuracy and 0.17\% in Top-5 accuracy). Notice that, the number of parameters of VGG-16 with DualConv does not change much on the ImageNet dataset. This is because the last fully connected (non-convolutional) layers occupy most of the parameters, \textit{i.e.}, about 102M on ImageNet dataset. In the case for ResNet-50, the model with DualConv significantly decreases the computational cost and parameters of the original ResNet-50 model with a slight drop in accuracy (0.18\% in Top-1 accuracy and 0.29\% in Top-5 accuracy) when $G$=2. Furthermore, from \tabref{table6}, we can see that our proposed DualConv achieves better accuracy than model compression methods and other efficient convolutional filters (\textit{i.e.}, GroupConv and HetConv) with similar parameters and computational cost.

\subsection{MobileNetV1 and MobileNetV2 on ImageNet}
We also experiment with the DualConv-modified MobileNetV1 and MobileNetV2 network architectures on the large-scale ImageNet dataset. The stride of the first depthwise separable convolutional layer in the original or DualConv-modified MobileNetV1 network is changed back to 2 for ImageNet dataset. The hyperparameter settings in MobileNetV1 are the same as those in ResNet-50 on ImageNet (as described in \secref{std_imagenet}).
The values of the hyperparamters in MobileNetV2 are set as: weight decay=4e-5 and initial learning rate=0.05. Besides, Cosine learning rate decay strategy is used, and the network is trained for 150 epochs. 

\tabref{table8} shows that although DualConv increases the accuracy of MobileNetV1 network (up to 1.64\% in Top-1 accuracy and 1.11\% in Top-5 accuracy), it also increases the network parameters and computational cost. As for MobileNetV2, the model parameters and computational cost are reduced by applying DualConv. However, the accuracy of MobileNetV2 with DualConv is lower than the original MobileNetV2. A possible reason for this is that we use DualConv to replace the entire inverted residual block in MobileNetV2 network, activating the feature maps once, while the inverted residual block activates the feature maps twice. 

Our results show that DualConv can be integrated in both standard and lightweight network architectures to increase the network accuracy and reduce the network parameters and computational cost. Our experiments also demonstrate that the proposed DualConv can fit to various image classification datasets well and has a strong generalization capability.

\begin{table}[t]
\caption{Performance of MobileNetV1 and MobileNetV2 with DualConv on ImageNet using different settings.}
\footnotesize
\begin{center}
\setlength{\tabcolsep}{0.65mm}{
\begin{tabular}{|l|c|c|c|c|c|c|}
\hline
Model&Top-1&Top-5&FLOPs&Params&GTime&CTime \\
 &Acc(\%)&Acc(\%)& & &(ms)&(ms) \\
\hline\hline
MobileNetV1&70.65&89.69&568M&4.23M&0.80&21.73\\
\hline
MobileNetV1\_GC\_G2&67.90&88.02&2.44G&15.17M&0.77&53.96\\
MobileNetV1\_HC\_P4&70.47&89.61&1.63G&10.46M&0.97&55.33\\
\hline
MobileNetV1\_G2&\textbf{72.29}&\textbf{90.80}&2.98G&18.31M&0.93&70.76\\
MobileNetV1\_G4&72.13&90.64&1.77G&11.24M&0.84&46.27\\
MobileNetV1\_G8&71.15&89.91&1.16G&7.71M&0.84&34.76\\
MobileNetV1\_G16&70.47&89.48&854.82M&5.94M&0.84&30.80\\
MobileNetV1\_G32&68.58&88.25&703.09M&5.06M&0.82&\textbf{30.77}\\
\hline\hline
MobileNetV2&69.22&88.93&300.79M&3.50M&0.80&20.62\\
\hline
MobileNetV2\_G2&\textbf{65.45}&\textbf{84.40}&221.46M&2.67M&0.80&\textbf{16.52}\\
MobileNetV2\_G4&62.78&84.39&171.79M&2.35M&0.79&16.57\\
MobileNetV2\_G8&62.20&84.12&146.95M&2.19M&0.79&17.30\\
MobileNetV2\_G16&61.60&83.67&135.55M&2.11M&0.80&17.40\\
MobileNetV2\_G32&55.97&79.20&135.26M&2.07M&0.79&17.36\\
\hline
\end{tabular}}
\end{center}
\label{table8}
\end{table}

\subsection{Classification Time on ImageNet}

As pointed out in~\cite{ma2018shufflenetv2}, FLOPs is an indirect network efficiency metric but speed is a direct metric. Therefore, not only the computational cost (FLOPs) and number of parameters are important for the evaluation of efficient convolutional filters, but also the inference time is an important evaluation metric. We measure the inference time per image for each network model. The GPU inference time is recorded on a single Nvidia Tesla V100 GPU, and the CPU time is recorded using a single thread on an Intel Core i7-8700 CPU. Since the image classification models generally run fast on V100 GPU and there exists synchronization overhead between GPU threads, we cannot see an obvious difference in GPU inference time between different models. Hence, we discuss about the CPU inference time for the evaluated models. It is demonstrated in \tabref{table6} that the proposed DualConv not only reduces the parameters and computational cost of VGG-16 and ResNet-50, but also reduces their CPU inference time by 52\% and 24\%, respectively. Moreover, \tabref{table8} shows that although DualConv increases the inference time of MobileNetV1, but it reduces the model parameters, computational cost and inference time of MobileNetV2.

Because GroupConv leads to higher memory access cost when the number of groups ($G$) is larger~\cite{ma2018shufflenetv2}, we can see from the CPU inference time in \tabref{table6} that the DualConv-modified VGG-16 and ResNet-50 networks achieve faster speeds when $G$=16 and $G$=8, but not when $G$=32.

\begin{figure*}[t]
\centering
\includegraphics[width=0.96\textwidth]{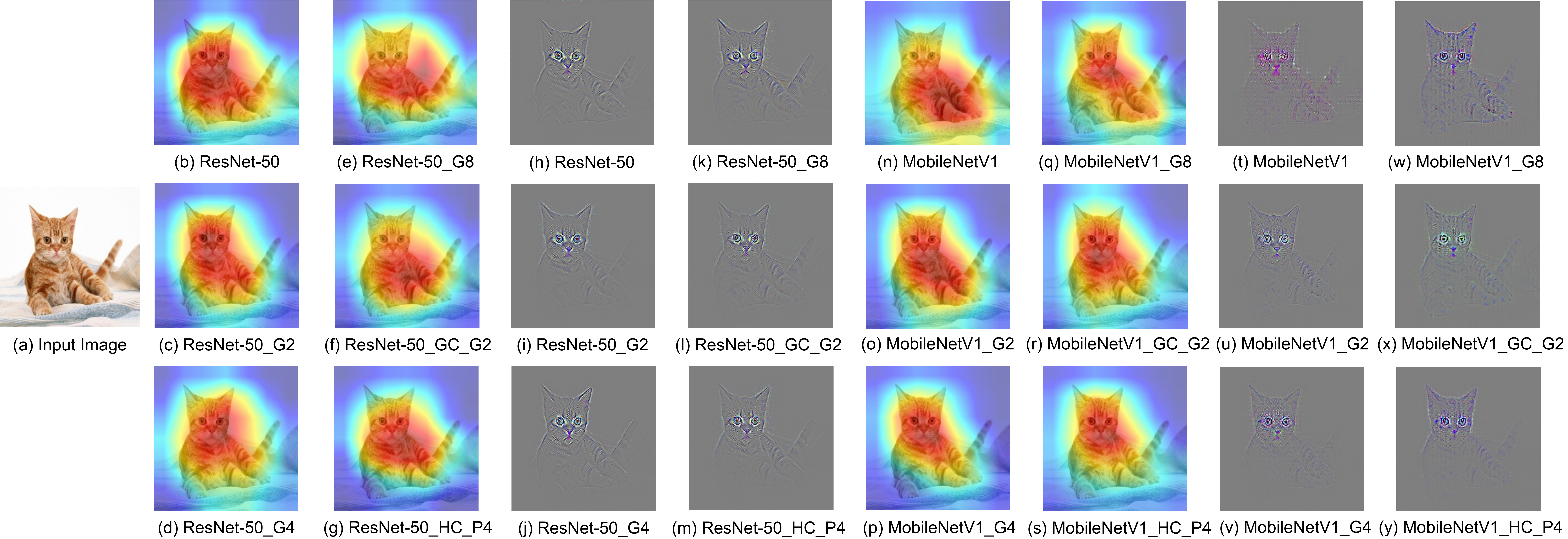} 
\caption{Visualizations of an example image for ResNet-50 and MobileNetV1 networks on ImageNet dataset. (a) is the original input image, (b)$\sim$(g) are the heatmaps obtained by Grad-CAM method on ResNet-50 networks, (h)$\sim$(m) are the guided Grad-CAM visualizations integrating guided backpropagation with Grad-CAM on ResNet-50 networks, (n)$\sim$(s) are the heatmaps obtained by Grad-CAM method on MobileNetV1 networks, and (t)$\sim$(y) are the guided Grad-CAM visualizations integrating guided backpropagation with Grad-CAM on MobileNetV1 networks.}
\label{fig3}
\end{figure*}

About the inference time comparisons with GroupConv and HetConv, we can see from \tabref{table6} and \tabref{table8} that when the values of $G$ (or $P$) 
are the same, GroupConv-modified models run faster than DualConv-modified models but HetConv-modified models run slower than DualConv-modified models (\textit{e.g.}, HetConv-modified ResNet-50 runs for 92.83ms while DualConv-modified ResNet-50 runs for 79.47ms). The former phenomenon is consistent with the number of FLOPs and parameters, but the latter is inconsistent because the HetConv filters are arranged in a shifted manner~\cite{singh2019hetconv}, which decreases the inference speed. This reflects the efficiency of the proposed DualConv.

\subsection{Visual Analysis}
To better illustrate the benefit of DualConv, we apply Grad-CAM~\cite{selvaraju2017grad} and guided backpropagation~\cite{springenberg2015striving} methods to visualize the ResNet-50 and MobileNetV1 networks on ImageNet dataset to obtain high-resolution class-discriminative visualizations. The resulting heatmaps and guided Grad-CAM visualizations of an example image are shown in \figref{fig3}. From \figref{fig3}, we can see that when DualConv ($G$=2) is applied to ResNet-50, the localization shown in Grad-CAM heatmap is more centred than other ResNet-50 networks, and the fine-grained details in its guided Grad-CAM visualization are clearer than other ResNet-50 networks except the original ResNet-50. Besides, DualConv-modified MobileNetV1 ($G$=2) presents the best localization and clearest fine-grained details among the compared MobileNetV1 networks.

\subsection{YOLO-V3 on PASCAL VOC}
To show that the proposed DualConv can generalize to different tasks, we apply DualConv to object detection model. YOLO-V3~\cite{redmon2018yolov3} is one of the common one-stage object detection frameworks using a single CNN to predict multiple bounding boxes and class probabilities.
Since the first convolutional layer of YOLO-V3 network is important for the low-level information extraction, it is not modified. All the remaining convolutional layers are modified with DualConv. The 3$\times$3 standard convolutions with stride 1 in YOLO-V3 are replaced by dual convolutions. However, we do not replace the 3$\times$3 convolutions with stride 2 in YOLO-V3 because the 1$\times$1 convolutions with stride 2 would harm the information preservation and channel fusion of input feature maps.

The original YOLO-V3 and DualConv-modified YOLO-V3 models are all trained from scratch. 
All the network models are trained for 100 epochs. We resize the input images to 416$\times$416 and use all 16551 images in the training and validation sets of PASCAL VOC 2007+2012 for training, and use the test set of PASCAL VOC 2007 for computing the mean average precision (mAP).
First, the average precision (AP) value of each class in PASCAL VOC 2007 is obtained by calculating the area under the precision-recall curve, and then these are averaged to get the mAP value. The inference time is recorded using one Nvidia Tesla V100 GPU.

From \tabref{table9}, we can see that the YOLO-V3\_G4 model which uses the proposed DualConv not only reduces the computational cost and the number of parameters by about 50\% but also improves the accuracy (mAP) by 4.4\% compared to the original YOLO-V3 model which uses 3$\times$3 standard convolutions. The inference time also improves from 26.66ms to 19.79ms (6.87ms faster). Therefore, DualConv not only compresses the model, but also improves the inference speed making it possible for object detection to be deployed on mobile platforms or embedded devices.

Note that the results in \tabref{table9} are obtained for networks using exactly the same settings, \textit{i.e.}, training from scratch on the same training set with the same number of epochs. Hence our comparison is fair. These results are not comparable to the YOLO-V3 model that is pre-trained on the ImageNet dataset which is much bigger.

\begin{table}[t]
\caption{Performance of YOLO-V3 with DualConv on PASCAL VOC 2007 test set.}
\footnotesize
\begin{center}
\begin{tabular}{|l|c|c|c|c|}
\hline
Model&mAP(\%)&FLOPs&Parameters&Time(ms)\\
\hline\hline
YOLO-V3&41.24&32.71G&61.63M&26.66\\
\hline
YOLO-V3\_G2&44.04&22.79G&42.41M&\textbf{18.17}\\
YOLO-V3\_G4&\textbf{45.64} &16.41G&30.05M&19.79\\
YOLO-V3\_G8&42.40&13.22G&23.88M&23.22\\
YOLO-V3\_G16&40.64&11.63G&20.79M&21.81\\
YOLO-V3\_G32&41.08&10.83G&19.24M&23.63\\
\hline
\end{tabular}
\end{center}
\label{table9}
\end{table}

\section{Conclusion}
We propose DualConv that combines 3$\times$3 group convolution with 1$\times$1 pointwise convolution solving the problem of cross-channel communication and preservation of the information in the original input feature maps. Compared to HetConv, DualConv improves network performance by adding minimal parameters. DualConv is applied to common network structures to perform image classification and object detection. By comparing the experimental results of standard convolution and DualConv, the effectiveness and efficiency of the proposed DualConv is demonstrated. As seen from the experimental results, DualConv can be integrated in both standard and lightweight network architectures to increase the network accuracy and reduce the network parameters, computational cost and inference time. We also demonstrate that DualConv can fit to various image datasets well and has a strong generalization capability. Future research work will focus on deployment on embedded devices to further prove the efficiency of DualConv in practical applications.

\bibliographystyle{IEEEtran}
{\footnotesize
\bibliography{DualConv_ref}}

\end{document}